\documentclass[final]{cvpr}

\usepackage{times}
\usepackage{epsfig}
\usepackage{graphicx}
\usepackage{amsmath}
\usepackage{amssymb}
\usepackage{subfloat}
\usepackage[caption=false]{subfig}

\newcommand{\ignore}[1]{}
\usepackage{booktabs}
\usepackage{xspace}
\usepackage{comment}
\usepackage{bm,upgreek}
\usepackage[normalem]{ulem}

\newcommand*{\affaddr}[1]{#1} % No op here. Customize it for different styles.
\newcommand*{\affmark}[1][*]{\textsuperscript{#1}}

\usepackage[pagebackref=true,breaklinks=true,colorlinks,bookmarks=false]{hyperref}

\def\cvprPaperID{****} % *** Enter the CVPR Paper ID here
\def\confYear{CVPR 2021}

\begin{document}

%%%%%%%%% TITLE
\title{Deep Video Inpainting Detection}

\author{Peng Zhou\affmark[1], Ning Yu\affmark[1], Zuxuan Wu\affmark[1], Larry S. Davis\affmark[1], Abhinav Shrivastava\affmark[1] and Ser-Nam Lim\affmark[2]\\
\affaddr{\affmark[1]University of Maryland, College Park \qquad}
\affaddr{\affmark[2]Facebook AI}
}

\maketitle

%%%%%%%%% ABSTRACT
\begin{abstract}
This paper studies video inpainting detection, which localizes an inpainted region in a video both spatially and temporally. In particular, we introduce VIDNet, Video Inpainting Detection Network, which contains a two-stream encoder-decoder architecture with attention module. To reveal artifacts encoded in compression, VIDNet additionally takes in Error Level Analysis frames to augment RGB frames, producing multimodal features at different levels with an encoder. Exploring spatial and temporal relationships, these features are further decoded by a Convolutional LSTM to predict masks of inpainted regions. In addition, when detecting whether a pixel is inpainted or not, we present a quad-directional local attention module that borrows information from its surrounding pixels from four directions. Extensive experiments are conducted to validate our approach. We demonstrate, among other things, that VIDNet not only outperforms by clear margins alternative inpainting detection methods but also generalizes well on novel videos that are unseen during training.
\end{abstract}

%%%%%%%%% BODY TEXT

\section{Introduction}
\begin{figure}[t]
\centering
   \includegraphics[width=0.48\textwidth]{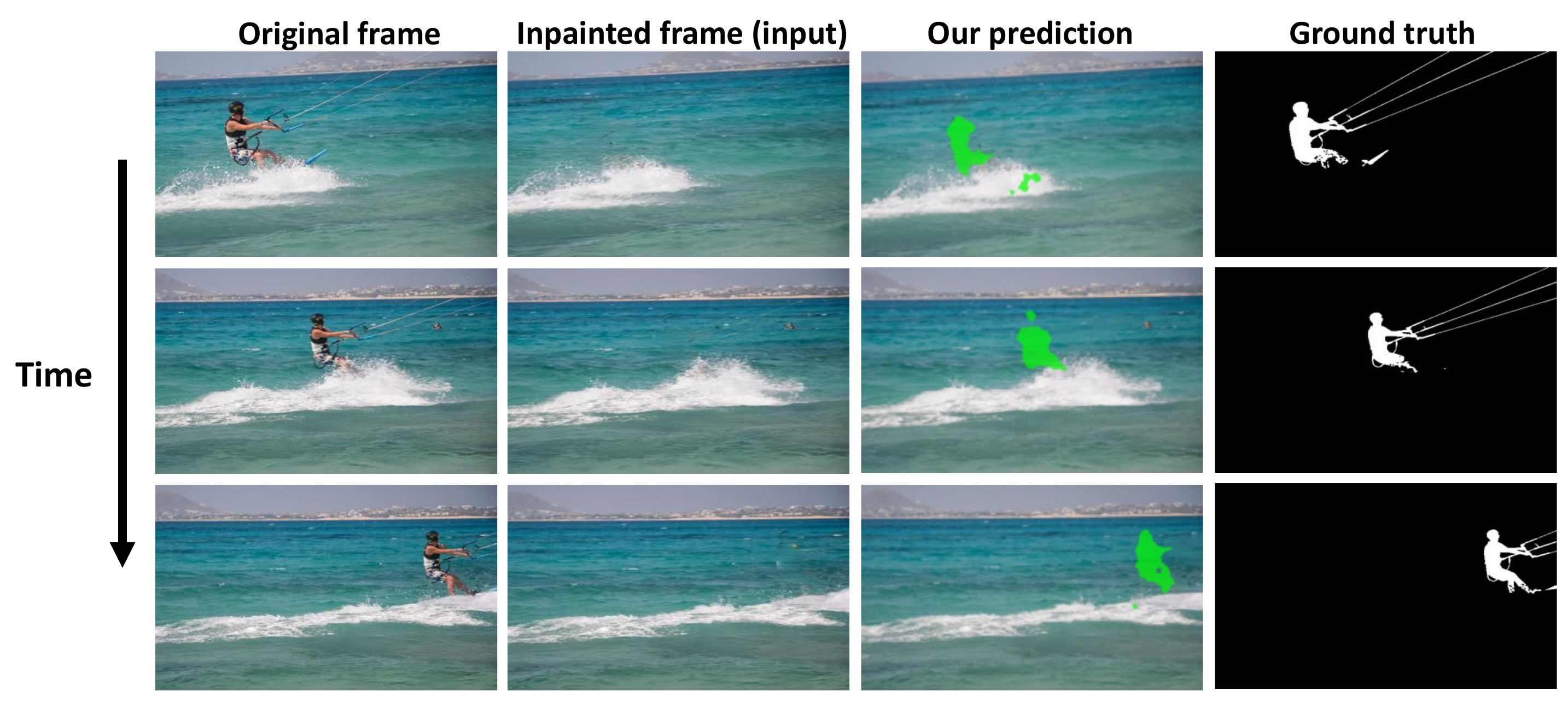}
   \vspace{-5pt}
   \caption{\textbf{Problem introduction}. Given an inpainted video (second column), we localize the inpainted region both spatially and temporally.}
\label{fig:concept}
\vspace{-15pt}
\end{figure}

Video inpainting, which completes corrupted or missing regions in a video sequence, has achieved impressive progress over the years~\cite{lee2019cpnet,kim2019dvi,xu2019fgvi,oh2019onion,chang2019free,huang2016temporally,yu2018generative,Xiong_2019_fgii,pathak2016context}. The ability to produce realistic videos that can be used in applications like video restoration, virtual reality, \etc, while appealing, brings significant security concerns at the same time since these techniques can also be used maliciously. By removing objects that could serve as evidence, malicious inpainting can result in serious legal and social implications including swaying a jury, accelerating the spread of misinformation on social platforms, \etc. Our goal in this work is to develop a framework for detecting inpainted videos constructed with state-of-the-art methods (see Fig.~\ref{fig:concept} for a conceptual overview). 

Although there are recent studies on detecting tampered regions in images~\cite{huh2018fighting,zhou2018learning,wu2019mantra,cozzolino2018forensictransfer}, very limited effort has been devoted to video inpainting detection. For image-based manipulation detection, existing approaches either focus on spliced regions or ``deepfake''-style face replacement instead of object removal based on inpainting. Additionally, most of them are designed specifically for images~\cite{li2019hpf,wu2008examplarinpainting} only and suffer from poor performance on videos. Learning robust video representations can help mitigate issues with single image detection.

% In addition, given a region to fill in, inpainting methods usually rely on

\begin{figure*}[t]
\centering
   \includegraphics[width=0.8\textwidth]{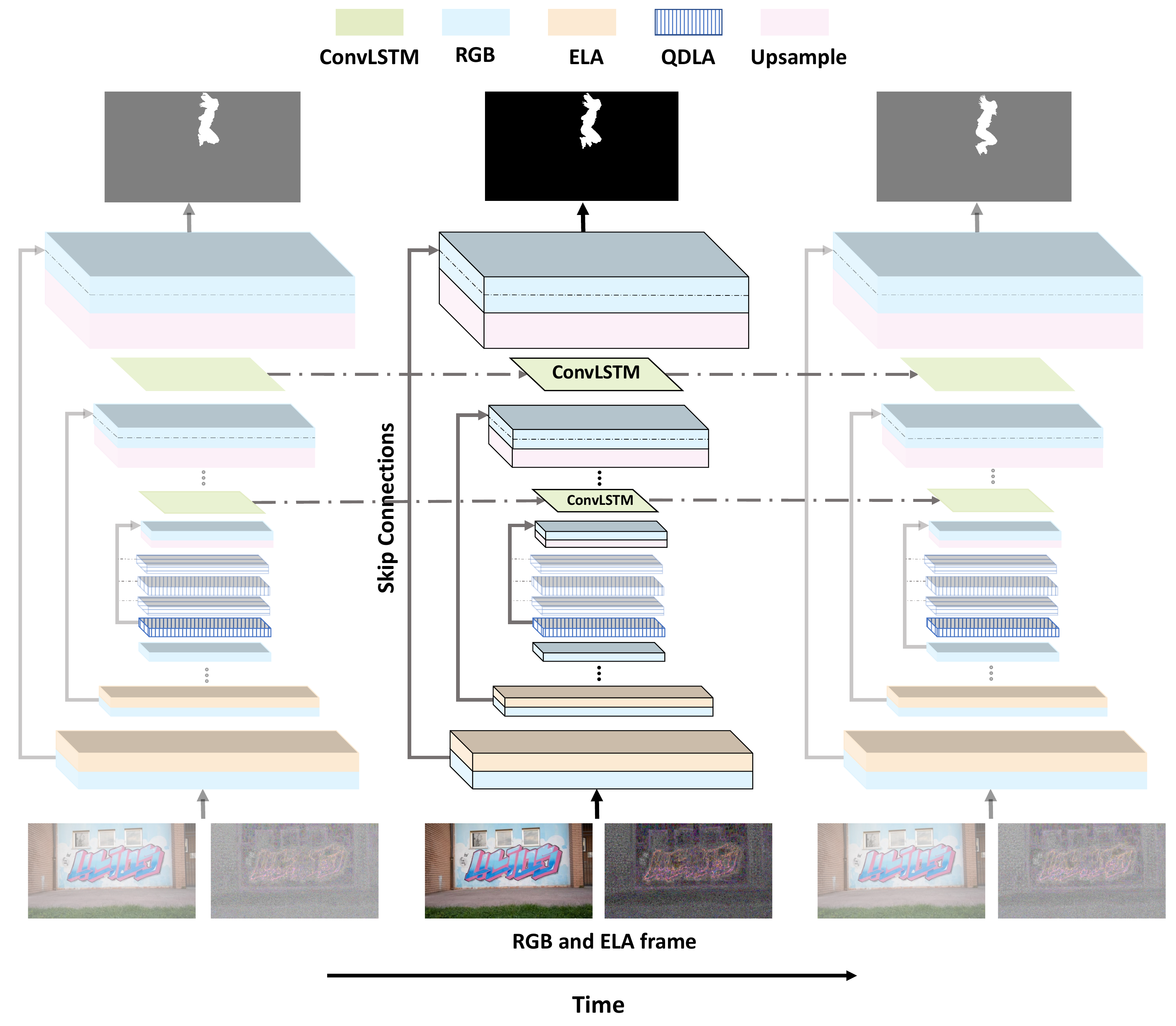}
   \vspace{-5pt}
  \caption{\textbf{Framework overview}. Given an RGB frame in a video, we first derive its corresponding ELA frame and compute multimodal features at different scales with both frames. We also introduce a quad-directional local attention module (striped) to the last encoded RGB features (colored blue) to explore spatial relationships among pixels from four directions. These encoded features are further input into a multi-layer ConvLSTM (colored green) for decoding, exploiting spatial and temporal relationships explicitly, to produce masks of inpainted regions. See texts for more details.}

\label{fig:frame}
\end{figure*}

In light of this, we introduce VIDNet, a video inpainting detection network, which is an encoder-decoder architecture with a quad-directional local attention module to predict inpainted regions in videos (as is shown in Fig.~\ref{fig:frame}). In particular, at each time step, VIDNet's encoder takes as input the current RGB frame, truncated from a pretrained VGG network~\cite{simonyan2014vgg}. Since videos are compressed based on discrete cosine transforms (DCT) and frames extracted are usually stored in JPEG format, we leverage ELA~\cite{wang2010ela} images as an additional input to the encoder to reveal artifacts like compression inconsistency (as is shown in Fig.~\ref{fig:ela}). We extract features from both ELA and RGB images with the encoder, producing five different multimodal features at different scales, that are further used jointly to train our inpainting detector.
In addition, given a missing region to fill in, inpainting methods leverage information from surrounding pixels of the region to make the region coherent spatially. Motivated by this, for RGB features from the last layer of the encoder, we introduce a quad-directional local attention module to attend to the neighbors of a pixel, allowing us to explicitly model spatial dependencies among different pixels during detection.

Finally, with multimodal features encoded at different scales, we leverage a four-layer Convolutional LSTM, serving as a decoder for inpainting detection. More specifically, the ConvLSTM at a certain layer not only takes in features from a previous time step but also features upsampled from a coarse level (\ie, a lower decoding layer). In this way both spatial relationships across different scales and temporal dynamics over time are leveraged to produce inpainted masks over time. The framework is trained end-to-end with backpropagation. We conduct experiments on the DAVIS 2016~\cite{perazzi2016davis} Dataset and the Free-form Video Inpainting Dataset~\cite{chang2019free}. VIDNet successfully detects inpainted regions under all different settings and outperforms by clear margins competing methods. We also show that VIDNet can be generalized to detect out-of-domain inpainted videos that are unseen during training.

Our contributions can be summarized as follows: 1) To the best of our knowledge, we introduce the first learning based approach for video inpainting detection. 2) We present an end-to-end framework for video inpainting detection, which models spatial and temporal relationships in videos. 3) We leverage multimodal features, \ie, RGB and ELA features, at different scales, for video inpainting detection. 4) We introduce a quad-directional local attention module to explicitly determine if a pixel is inpainted or not by attending to its neighbours.

\section{Related Work}

\textbf{Video Inpainting.} With the advance of recent image inpainting approaches~\cite{he2014imagecompletion,hays2007scene,iizuka2017globally,liu2012exemplar,pathak2016context,yu2018generative,liu2018partial,Xiong_2019_fgii,zhang2019internal}, more recent studies have investigated video inpainting. There are two lines of work --- patch based and learning based approaches. For patch based approaches, PatchMatch~\cite{barnes2009patchmatch} is a prominent approach which searches for similar patches in the surrounding region iteratively to complete the inpainted region. To achieve better quality, Huang \etal~\cite{huang2016temporally} explore an optimization based method to match patches and utilize information including color and flow as regularization. On the other hand, learning based approaches have been explored recently. Wang~\cite{wang2019video} propose a 3D encoder-decoder structure for video inpaining. Afterwards, Xu \etal~\cite{xu2019fgvi} leverages optical flow information to guide inpainting in videos in both forward and backward passes. Similarly, Kim \etal~\cite{kim2019dvi} estimate the proceeding flow as additional constraint while completing the missing regions. To maintain more frame pixels, Oh \etal~\cite{oh2019onion} use gated convolution to inpaint video frames gradually from the reference frame. Lee \etal~\cite{lee2019cpnet} copy and paste future frames to complete missing details in the current frame.
In contrast, our approach detects regions inpainted by these approaches.

\textbf{Manipulation Detection.} There are also approaches focusing on manipulation detection. Most mainly tackle splicing based manipulation and use clues specific to it \cite{ferrara2012cfa,cozzolino2016single,zampoglou2015web,cozzolino2018forensictransfer}. In particular, Zhou \etal~\cite{zhou2018learning} use both RGB and local noise to detect potential regions. Salloum \etal~\cite{salloum2018mfcn} rely on boundary artifacts to reveal manipulated regions in a multi-task learning fashion and Zhou \etal~\cite{zhou2018generate} improve its generalization ability with a generative model. Huh \etal~\cite{huh2018fighting} use meta-data to find inconsistent patches and Wu \etal~\cite{wu2019mantra} treat it as anomaly detection to learn features in a self-supervised manner. 

More related to our work are methods for image inpainting detection. ~\cite{wu2008examplarinpainting} is a classical approach that searches for similar patches matched by zero-connectivity. However, high false alarm  rates limit their applications in real scenarios. More recently, Zhu \etal~\cite{zhu201inpainting} use CNNs to localize inpainting patches within images. Li \etal~\cite{li2019hpf} explore High Pass Filtering (HPF) as the initialization of CNNs for the purpose of distinguishing high frequency noise of natural images from inpainted ones. However, the generalization and robustness is limited as these HPFs are learned given specific inpainting methods. In contrast, we combine both RGB information and ELA features as inputs to VIDNet, and show that our approach generalizes to different inpainting methods. In addition, without temporal guidance, the methods above cannot guarantee temporally consistent prediction like our approach.

\section{Approach}
VIDNet, Video Inpainting Detection Network, is an encoder-decoder architecture (See Fig.~\ref{fig:frame} for an overview the framework) operating on multimodal features to detect inpainted regions. In addition to RGB video frames, VIDNet utilizes Error Level Analysis frames (Sec.~\ref{encoder}) to identify artifacts incurred during the inpainting process. Motivated by the fact that inpainting methods typically borrow information from neighbouring pixels of the region to be inpainted, we introduce a multi-head local attention module (Sec.~\ref{attention}) which uses adjacent pixels to discover inpainting traces. Finally, we model the temporal relations among different frames with a ConvLSTM~(Sec.~\ref{decoder}). In the following, we describe the components of the model.

\subsection{Multimodal Features}\label{encoder}
 Learning a mapping directly from an inpainted RGB frame to a mask that encloses the removed object is challenging, since the RGB space is intentionally modified by replacing regions with their surrounding pixels to appear realistic. To mitigate this issue, we additionally augment RGB information with error level analysis features~\cite{wang2010ela} that are designed to reveal regions with inconsistent compression artifacts in compressed JPEG images. Note although videos are usually compressed in MPEG formats, extracted frames are often times stored in the format of JPEG. More formally, an ELA image is defined as:

\begin{equation}
   I_{ELA} = |I-I_{jpg}|,
\end{equation}
where $I_{ELA}$ is the ELA image, $I$ denotes the original image and $I_{jpg}$ denotes the recompressed JPEG image from the original image. 

\begin{figure}[t]
\centering
   \includegraphics[width=0.45\textwidth]{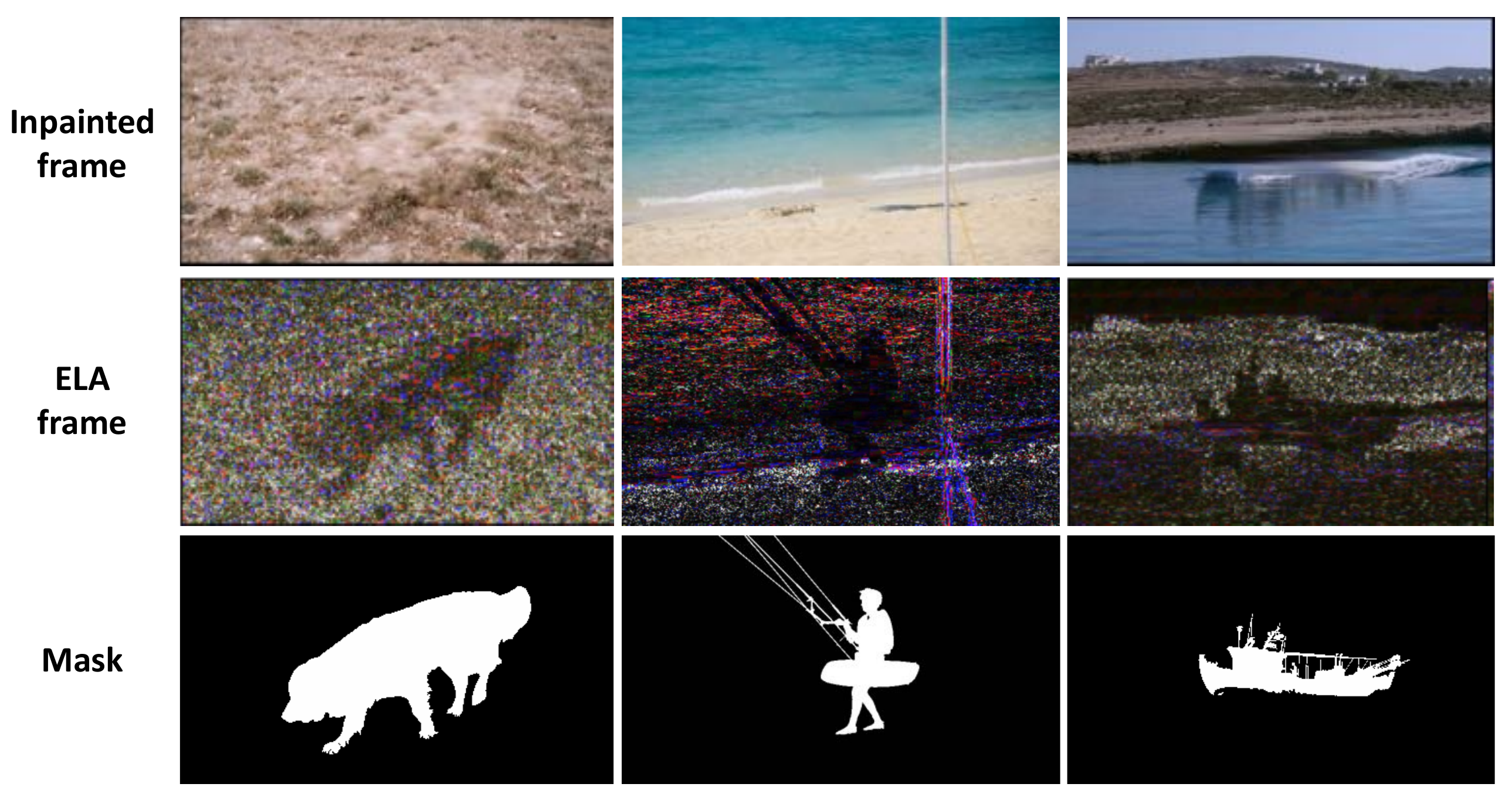}
   \vspace{-5pt}
   \caption{\textbf{ELA frame example}. From the top to the bottom: the inpainted RGB frame, its corresponding ELA frame, and the ground-truth inpainting mask. The inpainting artifacts, \eg, the dog, person and ship, stand out in ELA space while not easily seen in the RGB space.}
\label{fig:ela}
\vspace{-15pt}
\end{figure}

Fig.~\ref{fig:ela} illustrates the corresponding ELA images of sampled inpainted frames. Although ELA images have been used in forensics applications~\cite{zampoglou2015web,zampoglou2017large}, they tend to create false alarms when other artifacts like \eg, sharp boundaries, are present in the images, which requires ad-hoc judgement to determine whether a region is tampered. So, instead of only using ELA frames, we augment them with RGB frames as inputs to our encoder. (See results in Sec.~\ref{exp})

In particular, both the RGB and ELA frames are input to a two-stream encoder. Each stream, based on a VGG encoder, transforms the input image to high-level representations with five layers, yielding 5 feature representations at different scales. At each scale, we normalize the corresponding RGB and ELA features, respectively with $\ell_2$ normalization, and then apply one convolutional layer to absorb both features into a unified representation:

\begin{equation}
    f_l = \sigma(F(\;[\; f^{RGB}_l \;~|~ \; f^{ELA}_l  \;]))~~(l<5),
\end{equation}
where $[|]$ denotes feature concatenation, $f_l$ denotes the feature at $l$-th layer. $f^{RGB}_l$, $f^{ELA}_l$ denote the $L2$ normalized RGB and ELA features at layer $l$, respectively. $F$ represents the convolutional layer and $\sigma$ denotes the activation function. The fused representation at each level is further used for decoding. For $l=5$, we simply use RGB features as we find that high-level ELA features are not helpful.

% \subsection{Self-guided Recursive Filter Layer}
\subsection{Quad-Directional Local Attention}\label{attention}
Inpainting methods aim to replace a region with pixels from its surrounding areas for photorealistic visual effect. Therefore, when determining whether a pixel is inpainted or not, it is important to examine its surrounding pixels. Inspired by recursive filtering techniques that model pixel relations from four directions for edge-preserving smoothing, we introduce a quad-directional local attention module to explore spatial relations among adjacent pixels. 

We learn four attention maps for four directions, left-to-right, right-to-left, top-to-bottom, bottom-to-top, to determine how much information to leverage from the pixels in the corresponding direction based on each map. More specifically, we use $F_{\rightarrow}$, $F_{\leftarrow}$,  $F_{\uparrow}$ and $F_{\downarrow}$ to denote functions that derive attention maps for the left-to-right, right-to-left, top-to-bottom and bottom-to-top four directions. In the following, we consider the left-to-right direction for simplicity. Given features $f_5$ from the last layer of the RGB stream, we first transform the features with $F_{\rightarrow}$ to have the same dimension as $f_5$, and then compute an attention map $A_{\rightarrow}$:
\begin{align}
    % &f_i = (1-w_i)f_i + w_if_{i-1}, \\
    A_{\rightarrow} = \sigma(F_{\rightarrow}(f_5; W_{\rightarrow})),
\end{align}
where $W_{\rightarrow}$ denotes the weights for the convolutional kernel, and $\sigma$ is the sigmoid function to ensure the attentional weights at each pixel are in the range of $[0, 1]$. Then, for each pixel in the feature map, we obtain information from the surrounding pixels as:
\begin{equation}
    f_{5\rightarrow} [k] = (1-A_{\rightarrow} [k]) f_{5} [k] +  A_{\rightarrow} [k]  f_{5} [k-1],
\end{equation}
where $k$ denotes the location of the pixel. Since we are considering attention from the left-to-right direction, $k-1$ indicates the pixel to the left of $k$. The current value of pixel $k$ is updated with information from its neighboring pixel, and the weight to balance the contribution $A_{\rightarrow}$ is derived with convolution, which aggregates information from a small grid in the original features. As a result, we attend to a small local region to compute the refined representation. We can derive $f_{5\leftarrow}$, $f_{5\uparrow}$ and  $f_{5\downarrow}$ similarly, and thus we have four different refined representations.

Note that the quad-directional attention module is similar in spirit to recursive filtering. However, in standard recursive filtering, a weight matrix, in the form of an edge map~\cite{chen2018deeplab} or a weighted map~\cite{liu2016recursivefilter}, is used for the attention map $A$ to guide the filtering to restore images or smooth feature maps. In contrast, our filtering can be considered as a form of self-attention---we derive attention maps by modeling similarities in a local region with convolutions conditioned on input features and the resulting maps are in turn used to refine features, allowing pixels to borrow information by attending to its adjacent pixels. In addition, the motivation of our approach can be seen as the ``reverse'' process of recursive filtering---in recursive filtering, information from surrounding pixels is diffused to make local regions coherent, whereas we wish to detect inconsistent pixels by attending to a neighboring region.

Furthermore, we compute four refined feature maps for four directions in a parallel way conditioned on the same feature map. An alternative is to generate a single feature representation by sequentially performing attention in four directions, \emph{i.e.}, $f_{5\rightarrow}$ is used as inputs to generate $f_{5\leftarrow}$, and so on and so forth, as in~\cite{chen2018deeplab}. However, we find in Sec.~\ref{exp} that the parallel multi-head approach offers better results, possibly due to the disentanglement of different directions.

\begin{figure}[t]
\centering
   \includegraphics[width=0.45\textwidth]{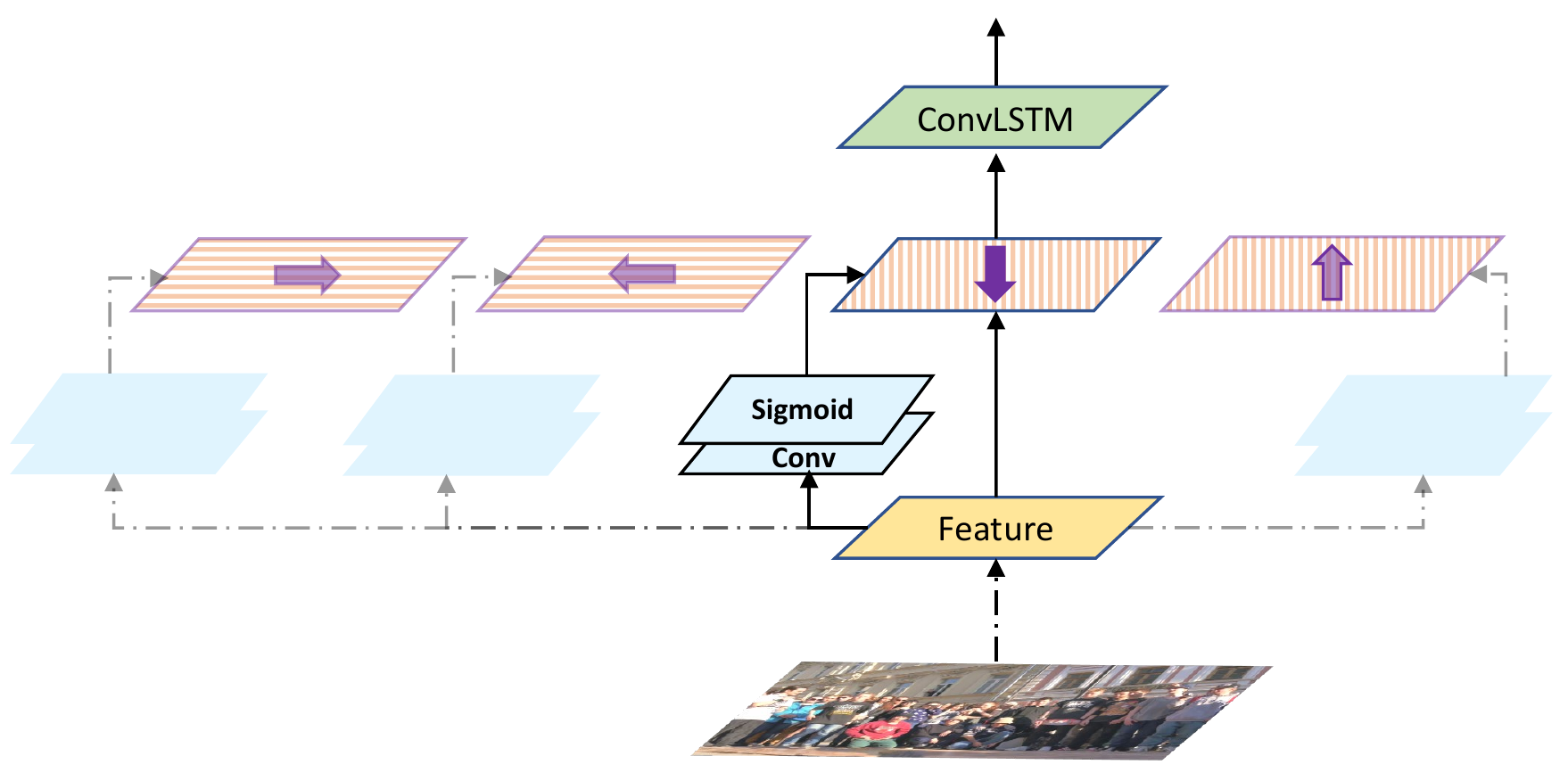}
   \vspace{-5pt}
   \caption{\textbf{The quad-directional local attention module}. Given RGB features from the last layer of the encoder, we derive attention maps with a quad-directional local attention module. To detect whether a pixel is inpainted or not, the module attends to its neighbors from four directions. 
   %\snl{[Somehow, I don't understand this figure -- in fact this looks like you also do temporal attention which I don't think you do? I think this figure looks more for section 3.3 for LSTM so maybe move it there]}
   }
\label{fig:srf}
\vspace{-15pt}
\end{figure}

\subsection{ConvLSTM Decoder}\label{decoder}

Temporal information like inconsistency in the inpainted region over time is an important cue for video inpainting detection. To explore temporal relationships among adjacent frames, we use multiple ConvLSTM decoding layers to take features from the encoders and produce predicted detection results, which enables message passing from previous frames. More specifically, the decoder contains four ConvLSTM layers to process features from different spatial scales.
At each time step, taking into account both spatial and temporal information, we concatenate the skipped connected feature of the current frame and the upsampled feature from a lower level, as the inputs to the current ConvLSTM layer. More formally, for the $t$-th time step, the $i$-th ($2<=i<=4$) ConvLSTM computes the hidden states and cell contents for the $t+1$-th time step as:

\begin{align}
\; h^{t+1}_i \;, c^{t+1}_i & =~\mathrm{ConvLSTM}_{i}( \; g_{i}^t \; , h^t_i \;, c^t_i),\label{eq:lstm}
 \\
g_{i}^t & = ~\;[\; U(h_{i-1}^t) \; | \; f_{6-i}^t \;],\label{eq:input}
\end{align}
where $h^t_i$ and $c^t_i$ denote the hidden states and cell states for the $i$-th ConvLSTM, respectively, and $U$ denotes the function for bilinearly upsampling, which maps the outputs from a lower-level ConvLSTM with smaller feature maps to have the same dimension as the current one. In addition, $f_{6-i}^t$ is the skip connected feature of the frame $t$ from the encoder. 

When $i=1$, the first layer of the ConvLSTM takes features from the last layer of the encoder, \emph{i.e.} $f_5$ as inputs. Recall that we obtain four refined features based on $f_5$ with our quad-directional local attention module to identify pixels that are inconsistent with its neighbours from four directions. Thus, we use these refined features as inputs to ConvLSTM$_{1}$. We input them into the LSTM in the order of $f_{5\rightarrow}$, $f_{5\leftarrow}$, $f_{5\uparrow}$ and  $f_{5\downarrow}$ to obtain all the four directional features. 

 At each time step, we compute $g_5^t$ with Eqn.~\ref{eq:input} to produce a prediction $p^t$ for each QDLA direction via one convolutional layer. Finally, to explore non-linear relations among these four directional outputs, we fuse them with one additional convolutional layer to form the final prediction. During training, we divide each video into N clips with equal clip length. To encourage more intersection with the binary ground truth mask, we use IoU score~\cite{ren2017iou} as our loss function which is formulated as:

 \begin{equation}
    L(p,y) \!\!= \!\! 1- \frac{\sum P*Y}{\sum (P\!\!+\!\!Y\!\!-\!\!P*Y\!\!)+\!\!\epsilon},
 \end{equation}
 where $P$ and $Y$ denote the prediction and the binary ground truth mask, respectively. $\epsilon$ denotes a small number to avoid zero division.

The loss is updated once the ConvLSTM decoder goes through a single video clip to collect temporal information. By exploring spatial and temporal information recurrently,  predictions of inpainted regions become more accurate.

\subsection{Implementation Details}
We use PyTorch for implementation. Our model is trained on a NVIDIA GeForce TITAN P6000. The input to the network is resized to $240 \times 427$. The length of our video clips is set to 3 frames during training. To extract ELA frames, we recompress the corresponding RGB frames by quality factor 50 and compute their difference. Our feature extraction backbone is VGG-16~\cite{simonyan2014vgg} for both RGB and ELA features. To increase the generalization ability, we add instance normalization~\cite{ulyanov2016instance} layers to the backbone. The encoder is initialized from VGG-16 model pretrained on ImageNet~\cite{deng2009imagenet} and the decoder is initialized by Xavier initialization~\cite{glorot2010xavier}. We concatenate both RGB and ELA features up to the penultimate encoding layer. Afterwards, the features are passed into one convolutional and normalization layer to reduce the dimension by half to reduce training parameters. The QDLA module is only added to the last encoder layer to extract directional feature information based on ablation results in Sec.~\ref{exp}. The decoder is a 4-layer ConvLSTM. We use Adam~\cite{kingma2014adam} optimizer with a fixed learning rate of $1 \times 10^{-4}$ for encoder and $1 \times 10^{-3}$ for decoder. The optimizer of the encoder and decoder network are updated in an alternating fashion. To avoid overfitting, weight decay with a factor of $5 \times 10^{-5}$ and $50\%$ dropout~\cite{srivastava2014dropout} are applied. Only random horizontal flipping augmentation is applied during training. We train the whole network end-to-end for 40 epochs with a batch size of 4.

\section{Experiment}\label{exp}
We compare VIDNet with approaches on manipulation/image inpainting detection in this section to show the advantages of our approach on video inpainting detection. We also analyze the robustness of our approach under different perturbations and show both quantitative and qualitative results.

\begin{table*}[t]
\small
\centering
\begin{tabular}{lccc|ccc|ccc}
\toprule
 \multicolumn{1}{l}{}
 
 % & \multicolumn{2}{c}{\textbf{VI}*}  & \multicolumn{2}{c}{\textbf{OP}*}  & \multicolumn{2}{c}{\textbf{CP}}   & \multicolumn{2}{c}{\textbf{VI}}  & \multicolumn{2}{c}{\textbf{OP}*}  & \multicolumn{2}{c}{\textbf{CP}*}   & \multicolumn{2}{c}{\textbf{VI}*}  & \multicolumn{2}{c}{\textbf{OP}}  & \multicolumn{2}{c}{\textbf{CP}*}   \\
  & \multicolumn{1}{c}{\textbf{VI}*}  & \multicolumn{1}{c}{\textbf{OP}*}  & \multicolumn{1}{c}{\textbf{CP}}   & \multicolumn{1}{c}{\textbf{VI}}  & \multicolumn{1}{c}{\textbf{OP}*}  & \multicolumn{1}{c}{\textbf{CP}*}   & \multicolumn{1}{c}{\textbf{VI}*}  & \multicolumn{1}{c}{\textbf{OP}}  & \multicolumn{1}{c}{\textbf{CP}*}   \\
 Methods & {IoU/F1}& {IoU/F1}& {IoU/F1} & {IoU/F1}& {IoU/F1}& {IoU/F1} & {IoU/F1}& {IoU/F1}& {IoU/F1}\\ 
 \midrule
NOI~\cite{mahdian2009using}  &0.08/0.14 &0.09/0.14 &0.07/ 0.13 &0.08/0.14 &0.09/0.14 &0.07/0.13 &0.08/0.14 &0.09/0.14 &0.07/ 0.13 \\
CFA~\cite{ferrara2012cfa}   &0.10/0.14 &0.08/0.14 & 0.08/0.12  &0.10/0.14 &0.08/0.14 & 0.08/0.12  &0.10/0.14 &0.08/0.14 & 0.08/0.12    \\
COSNet~\cite{lu2019see} &0.40/0.48 &0.31/0.38 &0.36/0.45 &0.28/0.37 &0.27/0.35 &0.38/0.46 &0.46/0.55 &0.14/0.26 & 0.44/0.53\\
HPF~\cite{li2019hpf} &0.46/0.57 &0.49/0.62 &0.46/0.58 &0.34/0.44 &0.41 /0.51 &0.68/0.77 & 0.55/0.67 &0.19/ 0.29 &0.69/0.80 \\
GSR-Net~\cite{zhou2018generate}& 0.57/0.69  &0.50/0.63 &0.51/0.63 &0.30 /0.43 &0.74/0.82&0.80/0.85 & 0.59 /0.70 &0.22/0.33 &0.70/0.77 \\
Ours RGB (baseline) &0.55/0.67 &0.46/0.58  &0.49/0.63 &0.31/0.42 &0.71 /0.77 &0.78/0.86 &0.58/0.69 &0.20/0.31 &0.70/0.82 \\
VIDNet-BN (ours) &\textbf{0.62}/\textbf{0.73} &\textbf{0.75}/ \textbf{0.83} &\textbf{0.67}/\textbf{0.78}&0.30/0.42 &\textbf{0.80}/\textbf{0.86} &\textbf{0.84}/\textbf{0.92}  &0.58 /0.70 &0.23/0.32 &0.75/0.85 \\
VIDNet-IN (ours) &0.59/0.70 &0.59/ 0.71 &0.57/0.69 &\textbf{0.39} /\textbf{0.49} &0.74/0.82 &0.81/0.87&\textbf{0.59}/ \textbf{0.71} &\textbf{0.25}/\textbf{0.34} &\textbf{0.76}/\textbf{0.85} \\
\bottomrule
\end{tabular}
\caption{\textbf{mean $IoU$ and $F_1$ score comparison on inpainted DAVIS.} The model is trained on VI and OP inpainting, OP and CP inpainting, and VI and CP inpainting respectively (denoted as `*').}
\label{tab:viop}
\end{table*}

\subsection{Experiment setup}

\textbf{Dataset and Evaluation Metrics.} Since DAVIS 2016~\cite{perazzi2016davis} is the most common benchmark for video inpainting, which consists of 30 videos for training and 20 videos for testing, we evaluate our approach on it for inpainting detection. We generate inpainted videos using SOTA video inpainting approaches --- VI~\cite{kim2019dvi}, OP~\cite{oh2019onion} and CP~\cite{lee2019cpnet}, with the ground truth object mask as reference. To show both the performance and generalization, we choose two out of the three inpainted DAVIS for training and testing, leaving one for additional testing. The training/testing split follows DAVIS default setting. We report the $F_1$ score and mean Intersection of Union (IoU) to the ground truth mask as evaluation metrics.

We compare our method with both video segmentation methods COSNet~\cite{lu2019see} and manipulation detection methods including \textit{NOI}~\cite{mahdian2009using}, \textit{CFA}~\cite{ferrara2012cfa}, \textit{HPF}~\cite{li2019hpf} and \textit{GSR-Net}~\cite{zhou2018generate}. Our baselines are shown below and see our \textbf{supplementary} for details on other approaches.

\textit{Ours RGB (baseline)}: Our baseline approach which feeds as input RGB frame only. No QDLA module is applied.

\textit{VIDNet-BN (ours)}: Our batch normalization~\cite{ioffe2015batch} version.

\textit{VIDNet-IN (ours)}: We report this as our main results, which replaces the batch normalization in encoder by instance normalization.

\subsection{Sanity Check}
Following~\cite{huh2018fighting}, we first check the ability of our learned model to distinguish between original and inpainted video frames. We compare models trained on VI and OP for simplicity. We add the original uninpainted videos to test sets for evaluation, and average the prediction score for every frame as frame-level score. Afterwards, we report the AUC classification performance in Tab.\ref{tab:check}. (Inpainted frames are labeled positive) Our model achieves better performance for all the three algorithms compared to other methods, indicating the advantages of our learned features to classify between inpainted and original videos.

\begin{table}[]
\centering
\begin{tabular}{p{3cm}ccc}
\hline
 Methods & \textbf{VI}* &\textbf{OP}* &\textbf{CP}  \\ 
 \hline
HPF~\cite{li2019hpf} & 0.718 &0.640 &0.845\\
GSR-Net~\cite{zhou2018generate} &0.762 &0.758 &0.834 \\
VIDNet-IN (ours) &\textbf{0.778} &\textbf{0.768} &\textbf{0.884} \\
\hline
\end{tabular}
\caption{\textbf{Sanity check for inpainting classification AUC comparison.} The results are tested on the three inpainting algorithms, and all the model are trained on VI and OP inpainted DAVIS.}
\label{tab:check}
\vspace{-15pt}
\end{table}

\subsection{Main Results}
Tab.~\ref{tab:viop} highlights our advantages over other methods. Video segmentation method COSNet captures the flow difference between adjacent frames to segment objects. In contrast, manipulation detection methods are learned to find tamper artifacts and thus yields better performance. For all the three settings, our IN version outperforms other approaches in both trained and untrained inpainting algorithms, showing the generalization of our approach. Additionally, we show clear improvement over our baseline, indicating the effectiveness of our proposed ELA feature and QDLA module. Comparing across different inpainting algorithms, the performance degrades on the untrained algorithms, indicating a domain shift between trained and untrained inpainting algorithms. However, benefiting from diverse features and more focus on proximity regions, our method still results in better generalization compared with other approaches. Finally, the results indicate that our BN version generally has better performance on the in-domain training inpainting algorithms while IN version shows better generalization on the cross-domain one. Therefore, we provide both results as a trade off between in-domain performance and generalization.

\begin{table}[t]
\small
\centering
\begin{tabular}{p{2.7cm}ccc}
\hline
 \multicolumn{1}{l}{}
  & \multicolumn{1}{c}{\textbf{VI}*}  & \multicolumn{1}{c}{\textbf{OP}*}  & \multicolumn{1}{c}{\textbf{CP}}   \\
  \hline
 Methods & {IoU/F1}& {IoU/F1}& {IoU/F1}\\ 
 \hline
Ours ELA &0.460/0.578 &0.509/0.631 &0.417/0.546\\ 
Ours RGB (baseline) &0.552/0.671 &0.456/0.580  &0.493/0.625 \\
Ours w/o QDLA &0.559/0.682 &0.557/0.681 &0.512/0.644  \\
Ours frame-by-frame &0.558/0.683 &0.566/0.688 &0.532/0.664 \\
Ours RF edge &0.540/0.661 &0.460/0.591 &0.555/0.670\\
QDLA both features &0.555/0.680 &0.580/0.700 &0.495/0.635\\
Ours w/o ELA &0.568/0.691 &0.465/0.595 &0.560/0.678 \\
QDLA all layers &0.570/0.693 &0.469/0.585 &0.564/0.682 \\
VIDNet-IN (ours) &0.585/0.704 &0.588/ 0.707 &0.565/0.685 \\
\hline
\end{tabular}
\caption{\textbf{Ablation analysis.} The model is trained on VI and OP inpainting algorithms (denoted as `*').}
\label{tab:ablation}
\vspace{-10pt}
\end{table}

  \begin{figure*}[t]
\begin{center}
   \subfloat[JPEG perturbation (VI*, OP*, CP)]{\includegraphics[width=0.8\textwidth]{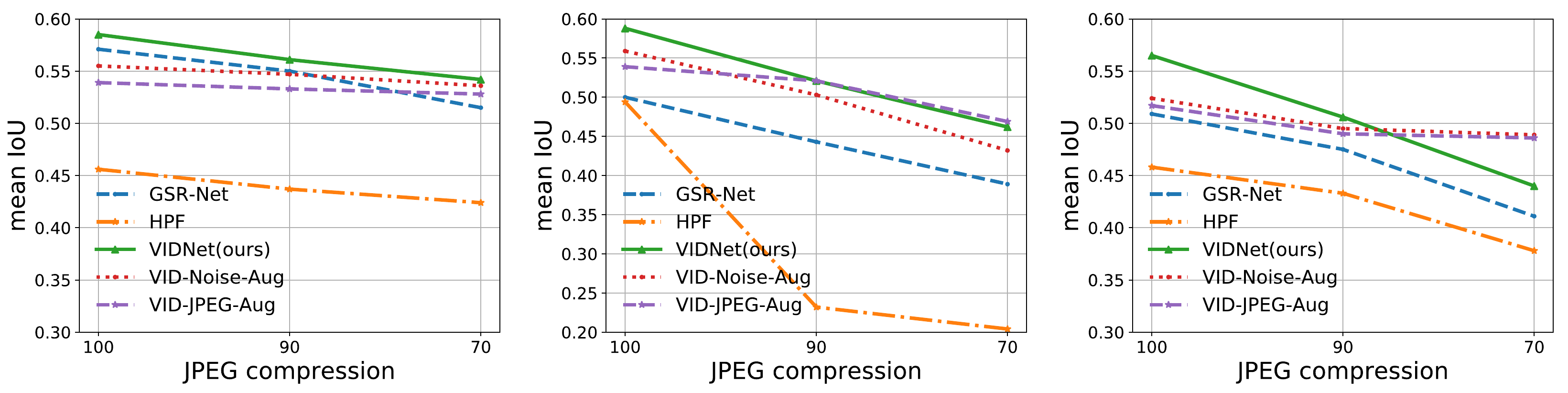}}
   \vspace{-5pt}
   \subfloat[Noise perturbation (VI*, OP*, CP)]{\includegraphics[width=0.8\textwidth]{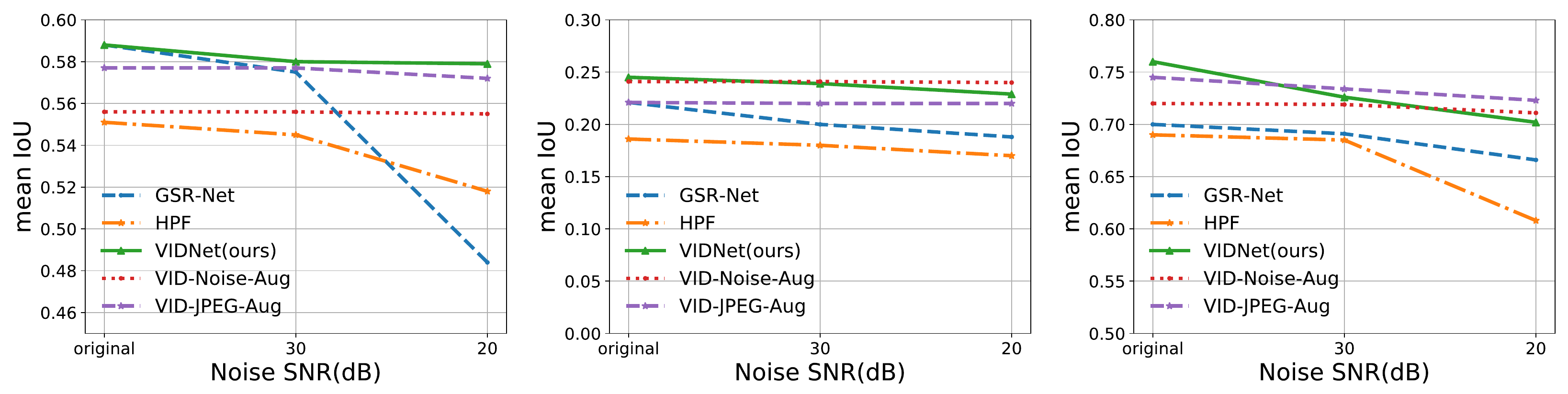}}
\end{center}
\vspace{-5pt}
   \caption{Mean IoU comparison under different perturbations. Perturbation in JPEG compression consists of the quality factor with 90 and 70; perturbation in noise consists of SNR 30dB and 20dB. Column from left to right is the result on VI, OP and CP inpainting. `*' denotes that the model is trained on these inpainting algorithms.}
   \vspace{-15pt}
   \label{fig:robust}
\end{figure*}

\subsection{Ablation Analysis}
We analyze the importance of each key component in our framework and the details are as follows:

\textit{Ours ELA}: The baseline architecture which only feeds ELA frame as input.

\textit{Ours w/o ELA}: Our full model without the ELA features.

\textit{Ours w/o QDLA}: Our full model without QDLA module.

\textit{Ours RF edge}: Similar to Chen \etal~\cite{chen2018deeplab}, we add additional edge branch and apply recursive filter to the final prediction. The output of edge branch is used as the reference to recursive filter layer. The loss function of the edge branch is a weighted binary cross entropy loss.

\textit{QDLA both features}: Our full model except that the input to QDLA module is the concatenation of both RGB and ELA feature from the $5$-th layer.

\textit{QDLA all layers}: Applying QDLA module to all the 5 encoding feature layers.

\textit{Ours frame-by-frame}: Instead of training with video clip length of 3, we train our full model frame-by-frame.

Tab.\ref{tab:ablation} displays the comparison results. Compared to baseline, the ELA feature alone yields worse performance. This perhaps because the ELA frame also contains other artifacts like sharp boundary, which leads to confusion without proper guidance from RGB contents. Adding QDLA module introduces feature adjacency relationship and thus leads to improvement. However, the higher features are more useful for our QDLA than lower ones when comparing to \textit{QDLA all layers}, and high level ELA features are less helpful than lower ones when comparing with \textit{QDLA both features}. Compared to \textit{Ours RF edge}, our QDLA module (\textit{Ours w/o ELA}) yields better performance because the boundary prediction degrades in video inpainting scenario and thus edge map contains false positives to guide the segmentation branch. In addition, the comparison between 
\textit{Ours frame-by-frame} and our final model verifies the importance of temporal information in video inpainting detection. Eventually, with QDLA module, ELA feature and temporal information, the performance gets boosted further. 

\begin{figure*}[t]
\begin{center}
   \includegraphics[height=0.7\textwidth]{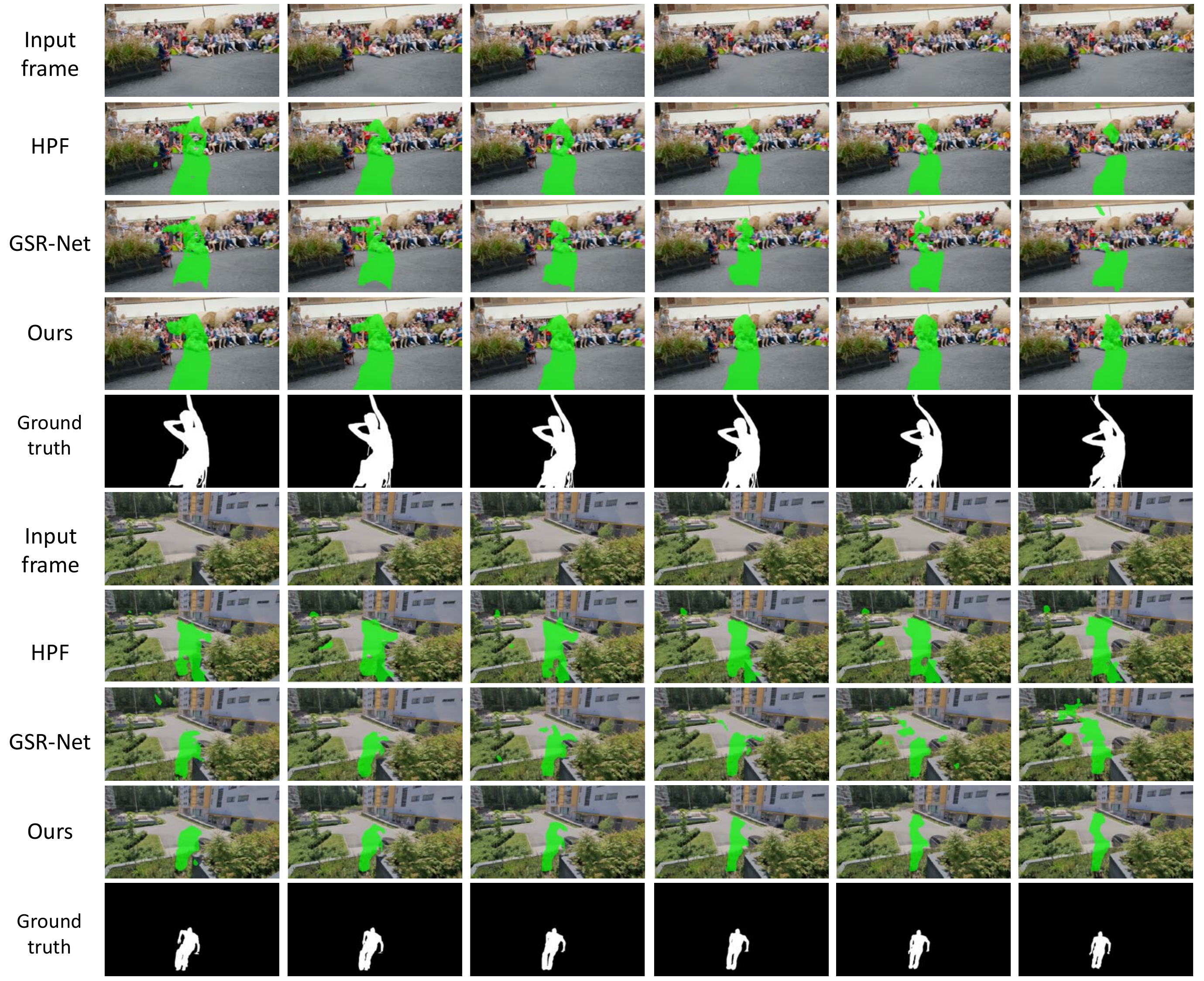}
\end{center}
\vspace{-10pt}
   \caption{Qualitative visualization on DAVIS. The first row shows the inpainted video frame. The second to fourth row indicates the final predictions from different methods. The fifth row is the ground truth. }
\label{fig:qr}
\vspace{-15pt}
\end{figure*}

\subsection{Robustness Analysis}
To test the robustness of our approach under noise and JPEG perturbation, we conduct experiments listed in Fig.~\ref{fig:robust}. We add Gaussian noise to the input frame with Signal-to-Noise Ratio (SNR) 30 and 20 dB and evaluate on these noisy frames, or recompress test frame with JPEG quality 90 and 70 for perturbation. Moreover, to study the effect of specific augmentation on performance, we apply noise and JPEG augmentation to our approach and make comparison together. The details of our augmentation is as follow.

\textit{VID-Noise-Aug}: Randomly apply Gaussian noise with SNR 20 dB to the input frames during training.

\textit{VID-JPEG-Aug}: Randomly apply JPEG compression with quality factor 90 to the input frames during training.

The robustness of our approach stands out under different perturbations. Compared to other approaches, HPF suffers more from perturbation because more high frequency noises will be introduced. With generative models for augmentation, GSR-Net shows good robustness. However, our approach outperforms GSR-Net as more modalities of video inpainting clues have been considered. Even though adding noise augmentation results in a small degradation on the initial performance, the robustness to both noise and JPEG perturbation has been improved. Similar observation is made on JPEG augmentation. See our \textbf{supplementary} for analysis under video compression perturbation.

\subsection{Results on Free-form Video Inpainting Dataset}
To further test the performance on different dataset, additional evaluation is provided on Free-form Video Inpainting dataset (FVI). FVI dataset~\cite{chang2019free} provides 100 test videos, which mostly targets multi-instance object removal. We directly apply their approach, which leverages 3D gated convolution encoder-decoder architecture for video inpainting, to generate the 100 inpainted videos. To test the generalization of our approach, we directly test the models trained on VI and OP inpainted DAVIS. 

Tab.~\ref{tab:fvi} displays the comparison results. Since both the dataset and inpainting approach are different, the performance degrades due to the domain shift. However, compared to other approaches, our method still achieves better generalization by a large margin. Also, compared with our baseline model which only uses RGB features, our approach shows clear improvement. This further validates the effectiveness to combine both RGB and ELA features and introduce spatial and temporal information for more evidence.

\subsection{Qualitative Results}
Fig.~\ref{fig:qr} illustrates the visualization of our predictions versus others under the same setting. Thanks to our ELA and RGB features which provide spatial clues, it is clear that our approach is able to obtain a closer prediction to the ground truth than other methods. Specifically, HPF only transfers RGB into noise domain, making it easier to produce false alarm. GSR-Net makes decision frame-by-frame, making the result less temporally consistent. In contrast, with the favor of temporal information, our prediction maintains temporal consistency. %However, the challenge lies in details and thin regions in that they have subtle inpainting artifacts. 

\begin{table}[t]
\small
\centering
\begin{tabular}{p{5cm}cc}
\hline
 \multicolumn{1}{l}{}
  & \multicolumn{1}{c}{\textbf{FVI}}   \\
  \hline
 Methods & {IoU/F1}\\ 
 \hline
NOI~\cite{mahdian2009using}  &0.062/0.107  \\
CFA~\cite{ferrara2012cfa}   &0.073/0.122    \\
HPF~\cite{li2019hpf} & 0.205/0.285 \\
GSR-Net~\cite{zhou2018generate} &0.195/0.288 \\
Ours RGB (baseline) &0.156/0.223 \\
VIDNet-IN (ours) &\textbf{0.257}/\textbf{0.367} \\
\hline
\end{tabular}
\caption{\textbf{Mean IoU and F1 score comparison on FVI.} The results are directly tested on FVI dataset, and all the model are trained on VI and OP inpainted DAVIS.}
\label{tab:fvi}
\vspace{-15pt}
\end{table}

\section{Conclusions}
We introduce learning based video inpainting detection in this paper. To reveal more inpainting artifacts from different domains, we propose to extract both RGB and ELA features and make concatenation. Additionally, we encourage learning from adjacent feature in a self-attended manner by introducing QDLA module. With both the adjacent spatial and temporal information, we make the final prediction through a ConvLSTM based decoder. Our experiments validate the effectiveness of our approach both in-domain and cross-domain. As shown in the results, there still exists a clear gap in the generalization and robustness, making the problem far from being solved. Involving some domain adaption strategies might be a remedy for this issue, which we leave for future research.

\section{Acknowledge}
We gratefully acknowledge support from Facebook AI and the DARPA MediFor program under cooperative agreement FA87501620191, ``Physical and Semantic Integrity Measures for Media Forensics''. 

{\small
\bibliographystyle{ieee_fullname}
\bibliography{egbib}
}

\end{document}